\documentclass[runningheads]{llncs}
\usepackage[T1]{fontenc}
\usepackage{microtype}
\usepackage{graphicx}
\usepackage{makecell}
\usepackage{booktabs}
\usepackage{amsmath}
\usepackage{float}
\usepackage{amsfonts}
\usepackage{bm}
\usepackage{subcaption}
\usepackage{algorithm}
\usepackage{algorithmic}
\usepackage{color}
\usepackage{hyperref}
\usepackage{xcolor}

\definecolor{darkblue}{RGB}{0,0,139}  
\def\vbeta{{\bm{\beta}}}

\usepackage{amssymb}
\usepackage{bbm}

\newcommand{\sR}{\mathbb{R}}
\newcommand{\AYA}{\tiny\texttt{AYA}}
\newcommand{\CARE}{\tiny\texttt{CARE}}
\newcommand{\EDGE}{\tiny\texttt{REL}}

\DeclareMathOperator*{\argmax}{arg\,max}

\def\Mood{{\texttt{Mood}}}

\def\Treat{\tiny{\texttt{Treat}}}

\def\AM{{\texttt{AM}}}
\def\PM{{\texttt{PM}}}

\newcount\comments  
\newcommand{\genComment}[2]{\ifnum\comments=1{\textcolor{#1}{\textsf{\footnotesize #2}}}\fi}

\usepackage{silence}
\definecolor{redorange}{RGB}{255,69,0}  
\definecolor{grassgreen}{RGB}{76, 153, 0}
\begin{document}
\title{Reinforcement Learning on Dyads to Enhance Medication Adherence}
%
%
\author{Ziping Xu\inst{1} \and
Hinal Jajal\inst{1} \and
Sung Won Choi\inst{2} \and
Inbal Nahum-Shani\inst{2} \and
Guy Shani\inst{2} \and
Alexandra M. Psihogios \inst{3} \and
Pei-Yao Hung \inst{2} \and
Susan A. Murphy\inst{1}}
\authorrunning{Xu et al.}
%
\institute{$^{1}$Harvard University, Cambridge, MA, USA \\
$^{2}$University of Michigan, Ann Arbor, MI, USA \\
$^{3}$Northwestern University, Feinberg School of Medicine, Chicago, IL, USA}
\maketitle              
\begin{abstract}



\footnotetext[1]{Corresponding author: Ziping Xu, \url{zipingxu@fas.harvard.edu}}

Medication adherence is critical for the recovery of adolescents and young adults (AYAs) who have undergone hematopoietic cell transplantation. However, maintaining adherence is challenging for AYAs after hospital discharge, who experience both individual (e.g. physical and emotional symptoms) and interpersonal barriers (e.g., relational difficulties with their care partner, who is often involved in medication management).
To optimize the effectiveness of a three-component digital intervention targeting both members of the dyad as well as their relationship, we propose a novel Multi-Agent Reinforcement Learning (MARL) approach to personalize the delivery of interventions.
By incorporating the domain knowledge, the  MARL framework, where each agent is responsible for the delivery of one intervention component, allows for faster learning compared with a flattened agent. Evaluation using a dyadic simulator environment, based on real clinical data, shows a significant improvement in medication adherence (approximately 3\%) compared to purely random intervention delivery. The effectiveness of this approach will be further evaluated in an upcoming trial.




\end{abstract}

\section{Introduction}

For patients who have 
undergone allogeneic \textbf{hematopoietic stem cell transplantation (HCT)}, strict adherence to medication regimens, such as prophylactic immunosuppressant therapy (i.e., calcineurin inhibitors, such as tacrolimus or cyclosporine, taken twice-daily), is crucial for mitigating the risk of acute graft-versus host disease (GVHD) \cite{gresch2017medication}. Acute GVHD occurs in 50-70\% of patients following HCT. A lower medication adherence (60\%) rate is shown to associate with higher severity of GVHD \cite{kirsch2014differences}.

The challenges of adherence management are amplified among \textbf{adolescents and young adults (AYAs)}, 
who often demonstrate poorer medication adherence \cite{psihogios2020needle,psihogios2022social,lyons2018theory}. For AYAs with cancer, self-management rarely involves the individual alone. Instead, up to 73\% of family care partners bear the primary responsibility for managing cancer-related medications for AYAs \cite{psihogios2020adherence}. 

Many of these dyads express a desire to move toward sharing these responsibilities with each other \cite{psihogios2020adherence}. Indeed, for AYAs with chronic health conditions, this developmental period often marks a shift from relying solely on a caregiver to taking more personal responsibility for health care. While shifts in autonomy versus dependence and navigating the ensuing family conflict that can arise from these new dynamics are normative parts of AYA development, difficult family interactions can have a detrimental impact on medication adherence. For example, in a meta-analysis \cite{psihogios2019family}, higher level of family conflict and lower levels of family cohesion were significantly associated with worse medication adherence across pediatric illnesses and age groups. 

After being discharged from the hospital, both individuals in the dyad face significant emotional and physical challenges as they adjust to managing medication regimens \textit{outside the hospital environment}.
For AYAs, the daily challenges of managing complex medication regimens, coping with treatment side effects, coping with stress, and maintaining normal activities in the context of a complex medical regimen can create distress in their home environment. Similarly, care partners must balance caregiving responsibilities with their personal obligations. Those who shoulder heavy caregiving responsibilities at home face higher physical and emotional stressors, which can impede their ability to provide effective care, make sound decisions, and support their AYA's  self-care \cite{reinhard2008supporting}.

This need for support outside the inpatient environment motivates the development of interventions that leverage \textbf{digital technologies} such as mobile devices \cite{uribe2023effectiveness}.
Digital interventions are promising for supporting both AYAs and care partners \textit{at home} on a daily basis, compared to traditional clinical support delivered with limited frequency (e.g., weekly clinical visits for post-HCT AYAs). There is strong heterogeneity across dyads and the users' context are constantly changing, which makes it important to personalize the intervention delivery to optimize the effectiveness of digital interventions.
Reinforcement Learning (RL), a machine learning technique that adaptively learns the optimal behavior in an unknown environment to maximize cumulative rewards, is a promising approach for achieving this personalization.
RL has been successfully applied in a variety of digital interventions \cite{liao2020personalized,battalio2021sense2stop,trella2024deployed,ghosh2024miwaves}.


In this paper, we describe our work in developing an RL algorithm for ADAPTS-HCT \cite{shani2024tips}. ADAPTS-HCT is a digital intervention for improving medication adherence by AYAs  over 100 days after receiving HCT. ADAPTS-HCT integrates three components: (1) twice-daily messages promoting positive emotions for the AYA, (2) daily messages focusing on coping and self-care strategies for the care partner, and (3) a weekly collaborative game for improving their relationship \cite{shani2024tips}. \textit{We call the three components AYA, care partner, and relationship component}, respectively. Table \ref{tab:interventions} summarizes these components. The fully developed intervention package will be evaluated in the upcoming clinical trial.



\begin{table}[h]
\centering
\caption{Intervention components in ADAPTS-HCT}
\begin{tabular}{c|c}
\hline
\textbf{Component} & \textbf{Intervention} \\
\hline
AYA & Twice-daily positive psychology messages \\
\hline
Care partner & Daily positive psychology messages \\
\hline
Relationship & \makecell[c]{Weekly collaborative game designed to facilitate\\ positive dyadic interpersonal relationship}\\
\hline
\end{tabular}
\label{tab:interventions}
\end{table}


\textbf{Goals.} Our goal is to design an RL algorithm that can personalize the delivery of these interventions to optimize their effectiveness. Given the complexity of the dyadic structure, we identify the following two key challenges:




\begin{enumerate}

\item \textbf{Managing multiple intervention decisions across different multi-scales.} There are three intervention components, each requiring decisions to be made at a different time scales. 
The decision-making occurs twice daily for AYAs, daily for care partners, and weekly for the relationship component. Making decisions on multiple timescales complicates the  algorithm design.
\item \textbf{Accelerating learning in noisy, data-limited settings.} 
Observed data in  digital intervention deployment is quite noisy \cite{trella2022designing}. Furthermore, limited data will be available to support in decision making for dyads recruited early in the clinical trial.  
Additionally, less data is available for learning  decisions that occur at slower timescales. These factors necessitate a sample-efficient algorithm that learns faster given limited data.

\end{enumerate}




\subsubsection{Contribution.} Our contribution is a novel  multi-agent RL (MARL) framework involving three RL agents, where each agent is responsible for making decisions for one specific intervention component and operates at the timescale corresponding to its intervention component timescale, which directly addresses challenge (1) about multi-scale decision-making.

The use of MARL decouples the decision processes of different intervention components, thus  improving  interpretability of the agent model design. This improved interpretability allows us to incorporate domain knowledge into the agent-specific algorithm designs to address challenge (2). To further accelerate learning, we propose a novel \textbf{reward engineering} method that learns a less noisy surrogate reward function for each component. Through evaluation in a carefully designed dyadic environment, we demonstrate both the superior performance of our proposed algorithm.




\section{RL Framework and Domain Knowledge}

We start with formulating the intervention decision making as an RL problem, where we underscore the challenge in the multiple time scales.
\begin{table}[pht]
    \centering
    \caption{Summary of variables about each target component}
    \label{tab:variables}
    \begin{tabular}{c|c|c|c}
    \toprule
    Target & Variable & Type & Description \\
    \midrule
    AYA & \makecell{$R_{w,d,t}^{\AYA}$ \\ $A_{w,d,t}^{\AYA}$ \\ $B_{w,d,t}^{\AYA}$} & \makecell{binary \\ binary \\ continuous} & \makecell{Medication adherence at time $t$ on day $d$ in week $w$ \\ Intervention at time $t$ on day $d$ in week $w$ \\ App burden at time $t$ on day $d$ in week $w$} \\
    \midrule
    Care partner & \makecell{$Y_{w,d}^{\CARE}$ \\ $A_{w,d}^{\CARE}$ \\ $B_{w,d}^{\CARE}$} & \makecell{continuous \\ binary \\ continuous} & \makecell{Psychological distress on day $d$ in week $w$ \\ Intervention on day $d$ in week $w$ \\ App burden on day $d$ in week $w$} \\
    \midrule
    Relationship & \makecell{$Y_w^{\EDGE}$ \\ $A_w^{\EDGE}$} & \makecell{binary \\ binary} & \makecell{Relationship quality at the end of week $w$ \\ Game intervention at the beginning of week $w$} \\
    \bottomrule
    \end{tabular}
    \end{table}
%
%
%
HCT treatment is followed by an outpatient 100 days (rounded to 14 weeks) twice-daily medication regimen. Decision times within the 14 weeks are denoted by $(w, d, t)$ where $w \in \{1, \dots, 14\}$ is the week index, $d \in \{1, \dots, 7\}$ is the day index, and $t \in \{1, 2\}$ is the decision window within a day. 

\textbf{Primary goal.} The primary goal is to make decisions at each decision time $t$ to maximize cumulative sum of medication adherence $\sum_{w=1}^{14}\sum_{d=1}^{7}\sum_{t=1}^{2}R_{w,d,t}^{\AYA}$, where $R_{w,d,t}^{\AYA}$ is medication adherence at window $t$ on day $d$ in week $w$. See Table \ref{tab:variables}, for selected information that will be collected on the dyad. 


\textbf{Action space.} All actions are binary (deliver versus do not deliver intervention content); see Tables~\ref{tab:interventions},\ref{tab:variables}. When the current time $(d = 1, t = 1)$ is the first decision time on the first day of the week, the agent chooses three actions corresponding to all three interventions components. If the current time is the first time  on a day after the first day of the week $(d> 1, t = 1)$, the agent chooses two actions corresponding to only the AYA intervention and the care partner components. At the second time on each day ($t=2$) the agent chooses one action corresponding to only the AYA intervention component. 

\textbf{Observation space.} Apart from the dynamic action space, we collect observations about different components at different time scales (Table \ref{tab:variables}). 


\vspace{-3mm}

\section{Domain Knowledge through Causal Diagram} 

\vspace{-3mm}

\begin{figure}[hpt]
\includegraphics[width=1.1\textwidth]{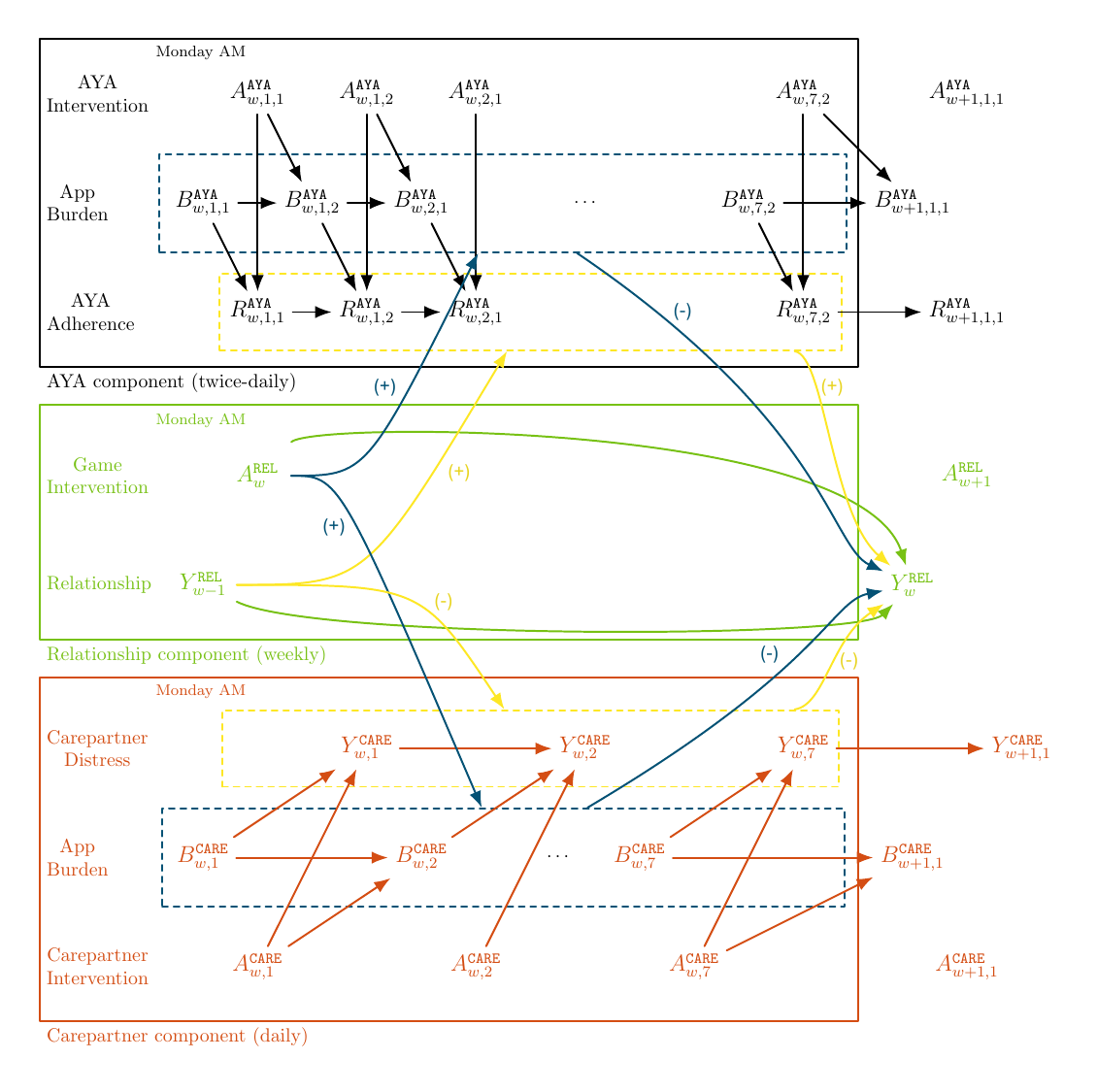}
    \captionsetup{parskip=0pt}
    \caption{Causal diagram for ADAPTS-HCT intervention \textsuperscript{1}. We categorize the variables into three components: AYA component (marked in \textbf{black}), care partner component (marked in \textcolor{redorange}{red}), and relationship component (marked in \textcolor{grassgreen}{green}). Each component operates at different time scales. Variables in the AYA component evolve on a twice-daily basis, while the care partner component operates on a daily basis. The relationship component operates on a weekly basis. The arrows indicate the direct causal effects.}
    \vspace{5mm}
    \small\textsuperscript{1} In the causal inference literature, this is called a causal Directed Acyclic Graph (DAG), a graphical representation of causal relationships among a set of variables \cite{pearl2000models}.
    \label{fig:dag}
\end{figure}



Our algorithm design is guided by domain knowledge encoded as the causal diagram in Fig. \ref{fig:dag}. 
This diagram describes the scientific team's understanding of the \textbf{primary causal relationships} between the variables in each component listed in Table \ref{tab:variables}. 

These variables are selected by the scientific team that are believed to capture the major causal pathways from our interventions to the primary outcome. Specifically, the app burdens are included as it has been well established in the mobile health literature that the app burden is a key moderator of the digital intervention through mobile health apps \cite{trella2022designing}. Literature has shown that a supporting positive framing in a dyad is associated with better adherence \cite{morrison2017medication,hoegy2019medication}, which motivates the inclusion of the relationship quality. The dyadic relationship is highly associated with the psychological distress of the care partner. The later is the major challenge faced by the care partner as they operate the caregiving role \cite{reinhard2008supporting}. 

We remark that the causal relationships are likely more complex and direct paths may exist between any two variables.  However the scientific team believes that these other paths are likely to be less detectable given the noise in digital intervention data.   We summarize the  primary pathways that interventions can take to effect the AYA's adherence in the following.  
\begin{enumerate}
\item \textbf{AYA intervention.} The AYA interventions $A_{w,d,t}^{\AYA}$ should directly influence the immediate AYA's adherence $R_{w,d,t}^{\AYA}$ (black arrows).

\item \textbf{Game intervention.} The game intervention $A_{w}^{\EDGE}$ has two pathways by which it is expected to effect  AYA's adherence. First, $A_{w}^{\EDGE}$ is expected to increase the AYA's burden $B_{w,d,t}^{\AYA}$ throughout the week $w$, which decreases the AYA's adherence $R_{w,d,t}^{\AYA}$ (\textcolor{darkblue}{blue arrows}). Second, the game intervention $A_{w}^{\EDGE}$ is expected to effect   next week AYA's adherence $R_{w+1,d,t}^{\AYA}$ by improving the end of the week relationship quality $Y_{w}^{\EDGE}$ (\textcolor{grassgreen}{green arrows}). 



\item 
\textbf{Care partner intervention.} The care-partner intervention $A_{w,d}^{\CARE}$ is expected to effect the AYA's adherence indirectly. First, $A_{w,d}^{\CARE}$ should decrease the care partner's psychological distress $Y_{w,d}^{\CARE}$. An decrease in psychological distress should increase the end of week relationship quality $Y_{w}^{\EDGE}$ (\textcolor{yellow}{yellow arrows}). Second, $A_{w,d}^{\CARE}$ should increase the care partner's burden $B_{w,d}^{\CARE}$, which should decrease the the end of week relationship quality $Y_{w}^{\EDGE}$ (\textcolor{darkblue}{blue arrows}). 

\end{enumerate}




\section{Proposed Multi-Agent RL Approach}

The conditional independence property observed from Fig. \ref{fig:dag} motivates us to design a multi-agent RL (MARL) comprising three decoupled agents: the AYA agent, the care partner agent, and the relationship agent. Each agent makes decisions at different time scales for their own component.


The MARL approach allows us to tailor the agent design choices for each agent to optimize the learning speed. Our base RL algorithm for each agent is Randomized Least Square Value Iteration (RLSVI) \cite{osband2016generalization}, which has been proven as stable in deployment of mobile health applications \cite{trella2024deployed,ghosh2024miwaves}. Additionally, we use linear models, which helps in discussions of the algorithm and its parameters with domain scientists. 

We construct agent-specific features based on Fig. \ref{fig:dag}. Specifically, the AYA agent's model uses its own variables ($B_{w, d, t}^{\AYA}, R_{w, d, t-1}^{\AYA}$) and the variables in the relationship component $(Y_{w-1}^{\EDGE}, A_{w}^{\EDGE})$. Similarly, the care partner agent uses its own variables, as well as the variables in the relationship component. The relationship agent's model uses $Y_{w-1}^{\EDGE}$, and previous weeks' $B_{w-1, 7, 2}^{\AYA}$, $B_{w-1, 7}^{\CARE}$, as well as a weighted average of AYA adherence and care partner distress in the past week.


\vspace{-3mm}

\subsection{Surrogate Reward Function Design Through Domain Knowledge}


The effects of care partner intervention and the game intervention are highly delayed. The game intervention improves end of week relationship quality with a significant delayed effect onto the adherence in the next week.
As shown in Fig. \ref{fig:dag}, the care partner intervention (positive messages for the care partner) is designed to mitigate the care partner's psychological distress, which only has indirect and delayed effects on the AYA's adherence. To address the this issue and accelerate learning, we engineer the reward function to account for the delayed effects. Similar reward engineering in the context of digital interventions is discussed in \cite{trella2023reward}. Our approach is distinct in that we explore the principles for incorporating domain knowledge to guide the reward design.

\textbf{Domain knowledge informed surrogate reward functions.} We introduce the surrogate reward functions for the relationship agent and the care partner agent. As informed by Fig. \ref{fig:dag}, the delayed effect of the game intervention is through the relationship quality and the AYA burden. 
This motivates us to fit a linear model to predict the sum of medication adherence within week $w$, $\sum_{d=1}^{7} \sum_{t=1}^{2} R_{w,d,t}^{\AYA}$, using $(1, Y_{w-1}^{\EDGE}, R_{w-1, 7, 2}^{\AYA}, {B}_{w, 1, 1}^{\AYA}, A_{w}^{\EDGE}, A_w^{\EDGE} \cdot Y_{w-1}^{\EDGE})$ as the covariates. 
To account for the delayed effect, we engineer the surrogate reward function for the relationship agent as:
$r_{w}^{\EDGE} = (1, Y_{w-1}^{\EDGE}, R_{w-1, 7, 2}^{\AYA}, {B}_{w, 1, 1}^{\AYA}, A_{w}^{\EDGE}, A_w^{\EDGE} \cdot Y_{w-1}^{\EDGE})\vbeta^{\EDGE}  
 + \max_{a}(1, Y_{w}^{\EDGE}, R_{w, 7, 2}^{\AYA}, {B}_{w+1, 1, 1}^{\AYA}, a, a \cdot Y_{w}^{\EDGE}) \vbeta^{\EDGE},$ where $\vbeta^{\EDGE} \in \sR^{5}$ are Bayesian linear regression estimates. 
The above reward yields a two-step greedy policy, which is a good enough approximation for the total sum of the medication adherence. We opt for a simple, linear model here because the bias trade-off is justified by the faster learning and reduction in noise. 

The design of the care partner agent is similar. A key observation is that the end of the week relationship quality blocks all the paths from the care partner variables to the next week' AYA adherence. Additionally, we believe that the relationship quality only have positive effects. Thus, we fit a linear model to predict the end of week relationship quality $Y_{w+1}^{\EDGE}$ using $(1, Y_{w,d}^{\CARE}, B_{w,d+1}^{\CARE}, Y_{w-1}^{\EDGE}, A_{w}^{\EDGE}, A_{w,d}^{\CARE})$ as covariates. The surrogate reward function is:
    $r_{w, d}^{\CARE} =  (1, Y_{w,d}^{\CARE}, B_{w,d+1}^{\CARE}, Y_{w-1}^{\EDGE}, A_{w}^{\EDGE}, A_{w,d}^{\CARE}) \vbeta^{\CARE}$, 
where $\vbeta^{\CARE} \in \sR^{5}$ are Bayesian linear regression estimates. 
\vspace{-3mm}
\section{Results}


\label{sec:testbed}


We simulate a \textit{dyadic environment} to evaluate the performance of the proposed framework. The environment design should replicate the noise level and structure that we expect to encounter in the forthcoming ADAPTS-HCT clinical trial. 

Our environment is based on Roadmap 2.0 dataset \cite{rozwadowski2020promoting} involving 171 dyads, each consisting of a patient undergone HCT (target person) and a care partner. Roadmap 2.0 provides daily positive psychology interventions to the care partner only. Roadmap 2.0 collects wearable devices data, for example, physical activity, and self-report data, for example, mood score.



We build upon previous work \cite{li2023dyadic} that also uses the Roadmap 2.0 data, but primarily focuses on AYA and relationship intervention components. We extend the environment to include the care partner intervention component. Specifically, we fit a separate multi-variate linear model for each component's outcome (ie., $R_{w,d,t}^{\AYA}, Y_{w,d}^{\CARE}, Y_w^{\EDGE}$) in the dataset. These models simulate the user trajectories under no intervention. A complete description and code of the dyadic environment is provided in supplementary material \footnote{\url{https://github.com/StatisticalReinforcementLearningLab/ADAPTS-HCT-AIME}}.


To simulate outcomes under treatments, we impute the treatment effects of the interventions and the effects of app burden, so the induced standard treatment effects (STE) \footnote{STE here is defined as the difference in the mean of the primary outcomes under the proposed intervention package and these under no intervention, which is further standardized by the standard deviation under no intervention. } are around 0.15, 0.3, and 0.5. These STEs are commonly seen in behavioral science studies \cite{cohen2013statistical}. 



\vspace{-3mm}

\subsection{Cumulative Adherence Improvement}





We simulate 25 dyads, the planned sample size in the upcoming pilot study, by sampling dyads sequentially with replacement from Roadmap 2.0 dataset. Each dyad is simulated for 14 weeks. We implement the following three algorithms: \texttt{SingleAgent}, \texttt{MultiAgent}, and \texttt{MultiAgent+SurrogateRwd}. Here the \texttt{SingleAgent} is the algorithm that trains a single agent that outputs all the three types of actions. The \texttt{MultiAgent} is the proposed MARL algorithm using the adherence as the learning reward signal for all three agents. The \texttt{MultiAgent+SurrogateRwd} is the proposed MARL algorithm using the surrogate reward functions. 

\begin{figure}[h]
    \centering
    \begin{subfigure}[b]{0.32\textwidth}
        \includegraphics[width=1\textwidth]{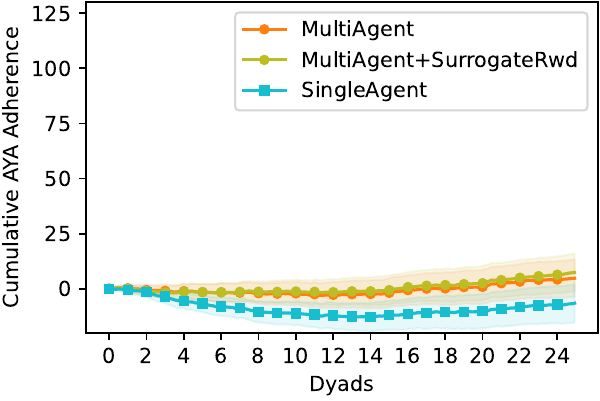}
        \caption{STE 0.15}
    \end{subfigure}
    \begin{subfigure}[b]{0.32\textwidth}
        \includegraphics[width=1\textwidth]{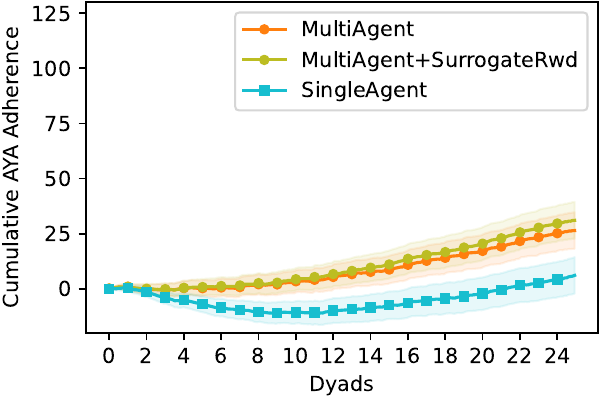}
        \caption{STE 0.3}
    \end{subfigure}
    \begin{subfigure}[b]{0.32\textwidth}

        \includegraphics[width=1\textwidth]{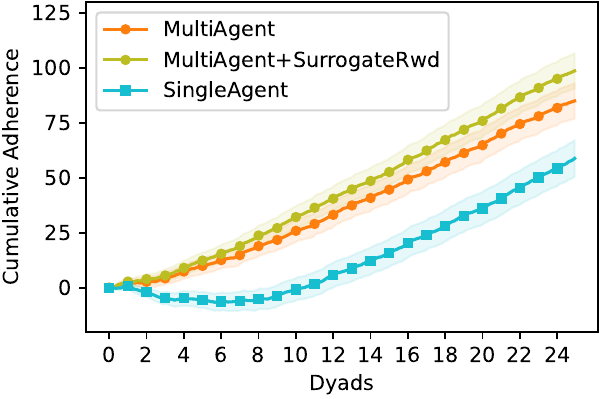}
        \caption{STE 0.5}
    \end{subfigure}
    \caption{Cumulative adherence improvement over the uniform random policy for all three components under dyadic environments with different STEs. The confidence interval is the standard deviation based on 1000 independent runs.} 
    \label{fig:cumulative_improvement}
    \vspace{-3mm}
\end{figure}

We report the cumulative adherence improvement over the purely random policy in Figure \ref{fig:cumulative_improvement}. We observe that all the algorithms can make more significant improvement over the random policy under a higher STE. \texttt{SingleAgent} takes longer to learn due to the larger number of parameters compared to \texttt{MultiAgent} as it does not fully utilize the dyadic structure and the causal relationship between the components described in Figure \ref{fig:dag}. We also see an advantage of using surrogate rewards through an increased cumulative adherence  at all levels of STE. Notice that for a low STE, the learning is slow, which is intuitive given that the the signal-to-noise ratio is low in such an environment. 
Additional ablation studies and analysis on the collaborating behavior is provided in the supplementary material. 

\vspace{-3mm}

\section{Discussion}


In this paper, we propose an MARL algorithm that effectively learns to optimize delivery of the ADAPTS-HCT digital interventions. While this presents a significant step towards preparing for the ADAPTS-HCT clinical trial, several challenges remain to be addressed. First, in the real-clinical trial, the participants are recruited incrementally with significant overlaps, whereas our dyadic environment assumes a simple sequential recruitment. 

\textbf{Acknowledgments.} This research was supported by various grants:  NIH/NIDA P50DA054039, NIH/NIBIB and OD P41EB028242, NIH/NIDCR UH3DE028723, and NIH/NIA 5P30AG073107-03 GY3 Pilots. Sung Won Choi is supported by NIH grants (K24HL156896 and R01CA249211 and UG1HL069330) and the Edith S Briskin/SKS Research Professorship and the Taubman medical Research Institute, which collectively support work related to GVHD.



\vspace{-3mm}

\bibliographystyle{splncs04}   
\bibliography{main.bib}

\newpage

\appendix

\section{Related Work}

Below we summarize the most relevant literature from both the medical lens and the algorithm lens.

\textbf{RL on social networks.} We design and implement RL on dyads that are small social networks in this paper. Existing works on RL on social networks are mostly focused on maximizing social influence or opinion spreading \cite{wang2021reinforcement,he2021reinforcement,yang2024balanced} with large scale social networks in mind. These problems are usually formulated as a constrained Markov Decision Process (CMDP) \cite{yang2024balanced}, where the goal is to allocate incentive to maximize the social influence or opinion spreading. Our focuses are on the challenges in the multi-scale decision making and the design of the RL algorithms that incorporate domain knowledge about the social networks. These differences make our algorithm designs unique contributions to the literature.

\textbf{Dyadic structure in health care.} Social relationships between patients and carepartners are proven to be important in many critical health outcomes. Studies have shown that the patient-caregiver dyad functions as a unit, with the well-being and coping strategies of one member significantly impacting the other \cite{shin2018supporting,mcpherson2024dyadic}. The quality of this relationship can affect treatment outcomes such as medication adherence \cite{psihogios2021understanding,kostalova2022medication,gresch2017medication}, and chronic disease management \cite{visintini2023medication,li2024usability}.

\textbf{Multi-agent RL (MARL).} Our proposed approach falls into the range of the independent learners in the MARL literature \cite{oroojlooy2023review}. Previous literature on MARL in a collaborative game focuses on finding the (approximate) Nash equilibrium of the game through interacting with an unknown environment \cite{wang2022cooperative,jin2021v}. However, in our paper, we emphasize the advantage of MARL in terms of its strong interpretability and being able to make decisions in multiple time-scales. Multi-agent RL has been commonly used in the cooperative game setting, for example, traffic signal control \cite{kolat2023multi,prabuchandran2014multi}. Our approach is novel in the use of surrogate reward functions for different agents to mitigate the delayed effect of the certain interventions. Our problem is also distinct from the typical cooperative game setting in its asymmetry in different agents' roles.

\section{Algorithm Details}
\label{app:algo}

We provide the complete details of the proposed \texttt{MultiAgent+SurrogateRwd} algorithm as well as the baseline \texttt{SingleAgent} algorithm.

We first introduce the infinite horizon RLSVI (Randomized Least Squares Value Iteration) algorithm in Alg. \ref{alg:base} \cite{russo2018tutorial}. This algorithm is a model-free posterior sampling approach that samples a random value function from its posterior distribution, and the agent acts greedily with respect to the sampled value function. We use the infinite horizon variant of RLSVI, which perturbs the Bayesian regression parameters with a random noise $\omega'$ (line 4). We introduce temporal correlation between the current noise $w'$ and the previous noise $w$ to introduce persistence in exploration.

\begin{algorithm}[H]
    \caption{Infinite Horizon RLSVI (Inf-RLSVI)}
        \begin{algorithmic}[1]
            \STATE{Input:} discount factor $\gamma \in \mathbb{R}$, previous dataset $\mathcal{D} = (s_i, a_i, r_i)_{i = 1}^{n-1} \cup \{s_{n}\}$, previous perturbation $w \in \mathbb{R}^d$, feature mapping $\phi: \mathcal{S} \times \mathcal{A} \mapsto \mathbb{R}^d$, previous parameter $\theta \in \mathbb{R}^{d}$
            \STATE Generate regression matrix and vector
            $$
                X \leftarrow\left[\begin{array}{c}
                \phi\left(s_1, a_1\right) \\
                \vdots \\
                \phi\left(s_{n-1}, a_{n-1}\right)
                \end{array}\right] \quad y \leftarrow\left[\begin{array}{c}
                r_1+\gamma \max _{\alpha \in \mathcal{A}} \langle \phi(s_2, \alpha), \theta \rangle \\
                \vdots \\
                r_{n-1}+\gamma \max _{\alpha \in \mathcal{A}}\langle \phi(s_{n}, \alpha), \theta \rangle
                \end{array}\right]
            $$
            \STATE Estimate value function
            $$
                \bar{\theta} \leftarrow \frac{1}{\sigma^2}\left(\frac{1}{\sigma^2} X^{\top} X+\lambda I\right)^{-1} X^{\top} y \quad \mathbf{\Sigma} \leftarrow\left(\frac{1}{\sigma^2} X^{\top} X+\lambda I\right)^{-1}
            $$
            \STATE Sample $w' \sim \mathcal{N}(\gamma w, (1-\gamma^2) \mathbf{\Sigma})$ and set $\theta' = \bar \theta + w'$
            \STATE \textbf{Output:} $\theta'$ and $w'$
        \end{algorithmic}
        \label{alg:base}
\end{algorithm}

We use the same hyperparameters $\lambda = 0.75$ and $\sigma = 0.5$ for all the algorithms, which achieves an overall good performance for all the algorithms.

\textbf{Additional notation.} We use $w, d, t$ to denote the week, day, and time of the decision. When we increment the time, we use $w, d, t+1$ to denote the next decisioin time right after $w, d, t$, and $w, d, t-1$ to denote the previous decision time right before $w, d, t$. Note that if $t = 1$, then $w, d, t-1$ is the evening decision time of the previous day.

\subsection{Single Agent Algorithm}

Our \texttt{SingleAgent} algorithm runs the RLSVI algorithm in Alg. \ref{alg:base} using the all the obervations available at time $w, d, t$ as the state variable:
$$
    S_{w, d, t} = 
    \left(Y_{w, d-1}^{\CARE}, Y_{w-1}^{\EDGE}, R_{w, d, t-1}^{\AYA}, \bar{Y}_{w-1}^{\AYA}, \bar{Y}_{w-1}^{\CARE}, B_{w, d, t}^{\AYA}, B_{w, d, t}^{\CARE}, A_{w, d}^{\CARE}, A_{w}^{\EDGE}\right) \in \mathbb{R}^{9}.
$$
Here we slightly abuse the notation by using $R_{w, d, t-1}^{\AYA}$ to represent the AYA adherence at half-day decision time prior to the current decision time $w, d, t$. This means that if $t = 1$, a morning decision time, then $R_{w, d, t-1}^{\AYA}$ is the AYA adherence at the previous night.

The \texttt{SingleAgent} algorithm has the three dimensional action space $\vec{a} = (a_1, a_2, a_3)^{\top} \in \{0, 1\}^{3}$, each entry corresponding to one of the three interventions. The second action $a_2$  will only be effective on a new day and the third action $a_3$ will only be effective on a new week. The feature mapping $\phi$ for the single agent algorithm is defined as
$$
    \phi(s, \vec{a}) = (1, s, a_1, a_2, a_3, s \cdot a_1, s \cdot a_2, s \cdot a_3) \in \mathbb{R}^{40}.
$$

\begin{algorithm}[hpt]
    \caption{\texttt{SingleAgent} Algorithm}
    \begin{algorithmic}[1]
        \STATE{Input:} discount factor $\gamma = 0.5$
        \STATE{Initialize:} $\theta_{1,1,1} = \mathbf{0} \in \mathbb{R}^{40}$; dataset $\mathcal{D}_{1,1,1} = \emptyset$
        \FOR{$w = 1, 2, \dots, 14$}
            \FOR{$d = 1, 2, \dots, 7$}
                \FOR{$t = 1, 2$}
                    \STATE{Call Algorithm \ref{alg:base} and update $\theta_{w,d,t}$}
                    \STATE{$\vec{a} = \argmax_{\alpha} \langle \phi(S_{w,d,t}, \alpha), \theta_{w,d,t} \rangle$}
                    \IF{$t = 1$ and $d = 1$ (New Week)}
                        \STATE{Set $A_{w}^{\EDGE} = \vec{a}_3$}
                    \ENDIF
                    \IF{$t = 1$ (New Day)}
                        \STATE{Set $A_{w,d}^{\CARE} = \vec{a}_2$}
                    \ENDIF
                    \STATE{Set $A_{w,d,t}^{\AYA} = \vec{a}_1$}
                    \STATE{Environment generates $R_{w,d,t}^{\AYA}$ and next state $S_{w,d,t+1}$}
                    \STATE{Update $\mathcal{D}_{w,d,t} = \mathcal{D}_{w,d,t-1} \cup \{(S_{w,d,t}, \vec{a}, R_{w,d,t}^{\AYA})\}$}
                \ENDFOR
            \ENDFOR
        \ENDFOR
    \end{algorithmic}
    \label{alg:single_agent}
\end{algorithm}

\subsection{MultiAgent Algorithm}

The \texttt{MultiAgent} algorithm runs an RLSVI agent for each of the three interventions. We use agent-specific feature mapping $\phi^{\AYA}, \phi^{\CARE}, \phi^{\EDGE}$ for the AYA, carepartner, and relationship agents, respectively. The state construction and the feature mapping for Q-value function are given by Table \ref{tab:state_feature}. The \texttt{MultiAgent} algorithm is described in Alg. \ref{alg:multi_agent}, where the carepartner and the relationship agents learns based on the naive rewards that are the sum of the AYA rewards over the day, and over the week, respectively (line 15 and line 18).

\begin{table}[hpt]
    \centering
    \caption{State and feature construction for the Q-value function by agent.}
    \label{tab:state_feature}
    \begin{tabular}{l|l}
        \toprule
        Agent & State or Feature Mapping \\
        \midrule
        AYA State & $S_{w, d, t}^{\AYA} = \left(R_{w, d, t-1}^{\AYA}, B_{w, d, t}^{\AYA}, Y_{w}^{\EDGE}, A_{w}^{\EDGE}\right) \in \mathbb{R}^{4}$ \\
        AYA Feature & $\phi^{\AYA}(s, a) = (1, s, a, s \cdot a) \in \mathbb{R}^{10}$ \\
        \midrule
        Carepartner State & $S_{w, d}^{\CARE} = \left(Y_{w, d-1}^{\CARE}, B_{w, d}^{\CARE}, Y_{w}^{\EDGE}, A_{w}^{\EDGE}\right) \in \mathbb{R}^{4}$ \\ 
        Carepartner Feature & $\phi^{\CARE}(s, a) = (1, s, a, s \cdot a) \in \mathbb{R}^{10}$ \\
        \midrule
        Relationship State & $S_{w}^{\EDGE} = \left(Y_{w-1}^{\EDGE}, B_{w, 1, 1}^{\AYA}, B_{w, 1}^{\CARE}, \bar{Y}_{w-1}^{\AYA}, \bar{Y}_{w-1}^{\CARE}\right) \in \mathbb{R}^{5}$ \\
        Relationship Feature & $\phi^{\EDGE}(s, a) = (1, s, a, s \cdot a) \in \mathbb{R}^{12}$ \\
        \bottomrule
    \end{tabular}
\end{table}

\begin{algorithm}[hpt]
    \caption{\texttt{MultiAgent} Algorithm}
    \begin{algorithmic}[1]
        \STATE{Input:} discount factor $\gamma^{\AYA} = 0.5$, $\gamma^{\CARE} = 0.5$, $\gamma^{\EDGE} = 0$
        \STATE{Initialize:} $\theta^{\AYA}_{1,1,1} = \mathbf{0} \in \mathbb{R}^{10}$; $\theta^{\CARE}_{1,1} = \mathbf{0} \in \mathbb{R}^{10}$; $\theta^{\EDGE}_{1} = \mathbf{0} \in \mathbb{R}^{12}$; dataset $\mathcal{D}_{1,1,1}^{\AYA} = \emptyset$; $\mathcal{D}_{1,1}^{\CARE} = \emptyset$; $\mathcal{D}_{1}^{\EDGE} = \emptyset$
        \FOR{$w = 1, 2, \dots, 14$}
            \STATE{Call Algorithm \ref{alg:base} using $\mathcal{D}_{w}^{\EDGE}, \gamma^{\EDGE}$, and update $\theta_{w}^{\EDGE}$}
            \STATE{Set $A_{w}^{\EDGE} = \argmax_{\alpha} \langle \phi^{\EDGE}(S_{w}^{\EDGE}, \alpha), \theta_{w}^{\EDGE} \rangle$}
            \FOR{$d = 1, 2, \dots, 7$}
                \STATE{Call Algorithm \ref{alg:base} using $\mathcal{D}_{w,d}^{\CARE}, \gamma^{\CARE}$, and update $\theta_{w,d}^{\CARE}$}
                \STATE{Set $A_{w,d}^{\CARE} = \argmax_{\alpha} \langle \phi^{\CARE}(S_{w,d}^{\CARE}, \alpha), \theta_{w,d}^{\CARE} \rangle$}
                \FOR{$t = 1, 2$}
                    \STATE{Call Algorithm \ref{alg:base} using $\mathcal{D}_{w,d,t}^{\AYA}, \gamma^{\AYA}$, and update $\theta_{w,d,t}^{\AYA}$}
                    \STATE{$A_{w,d,t}^{\AYA} = \argmax_{\alpha} \langle \phi^{\AYA}(S_{w,d,t}^{\AYA}, \alpha), \theta_{w,d,t}^{\AYA} \rangle$}
                    \STATE{Environment generates $R_{w,d,t}^{\AYA}$ and next state $S_{w,d,t+1}$}
                    \STATE{Update $\mathcal{D}_{w,d,t}^{\AYA} = \mathcal{D}_{w,d,t-1}^{\AYA} \cup \{(S_{w,d,t}^{\AYA}, A_{w,d,t}^{\AYA}, R_{w,d,t}^{\AYA})\}$}
                \ENDFOR
                \STATE{Compute care-partner reward $R_{w,d}^{\CARE} = \sum_{t = 1}^{2} R_{w,d,t}^{\AYA} / 2$} 
                \STATE{Update $\mathcal{D}_{w,d}^{\CARE} = \mathcal{D}_{w,d-1}^{\CARE} \cup \{(S_{w,d}^{\CARE}, A_{w,d}^{\CARE}, R_{w,d}^{\CARE})\}$}
            \ENDFOR
            \STATE{Compute relationship reward $R_{w}^{\EDGE} = \sum_{d = 1}^{7} R_{w,d}^{\CARE} / 7$}
            \STATE{Update $\mathcal{D}_{w}^{\EDGE} = \mathcal{D}_{w-1}^{\EDGE} \cup \{(S_{w}^{\EDGE}, A_{w}^{\EDGE}, R_{w}^{\EDGE})\}$}
        \ENDFOR
    \end{algorithmic}
    \label{alg:multi_agent}
\end{algorithm}

The \texttt{MultiAgent+SurrogateRwd} algorithm is described in Alg. \ref{alg:multi_agent_surrogate}. The only difference between the \texttt{MultiAgent} and \texttt{MultiAgent+SurrogateRwd} is that the later agent optimizes the surrogate reward functions, defined in Equ. (\ref{equ:game_rwd_app}) and Equ. (\ref{equ:care_rwd_app}), where the coefficients are estimated using Bayesian Ridge Regression, with the prior mean given in Table \ref{tab:prior}.

\begin{align}
    r_w^{\EDGE} = & (1, Y_{w-1}^{\EDGE}, {B}_{w, 1, 1}^{\AYA}, A_{w}^{\EDGE}, A_w^{\EDGE} \cdot Y_{w-1}^{\EDGE})\vbeta^{\EDGE} \nonumber \\
     &+ \max_{a \in \{0, 1\}}(1, Y_{w}^{\EDGE}, {B}_{w+1, 1, 1}^{\AYA}, a, a \cdot Y_{w}^{\EDGE}) \vbeta^{\EDGE}, \label{equ:game_rwd_app}
\end{align}

\begin{align}
        r_{w, d}^{\CARE} =  (1, Y_{w,d}^{\CARE}, B_{w,d+1}^{\CARE}, Y_{w-1}^{\EDGE}, A_{w,d}^{\CARE}) \vbeta^{\CARE}, \label{equ:care_rwd_app}
\end{align}

\begin{algorithm}[hpt]
    \caption{\texttt{MultiAgent+SurrogateRwd} Algorithm}
    \begin{algorithmic}[1]
        \STATE{Input:} discount factor $\gamma^{\AYA} = 0.5$, $\gamma^{\CARE} = 0.5$, $\gamma^{\EDGE} = 0$
        \STATE{Initialize:} $\theta^{\AYA}_{1,1,1} = \mathbf{0} \in \mathbb{R}^{10}$; $\theta^{\CARE}_{1,1} = \mathbf{0} \in \mathbb{R}^{10}$; $\theta^{\EDGE}_{1} = \mathbf{0} \in \mathbb{R}^{12}$; dataset $\mathcal{D}_{1,1,1}^{\AYA} = \emptyset$; $\mathcal{D}_{1,1}^{\CARE} = \emptyset$; $\mathcal{D}_{1}^{\EDGE} = \emptyset$
        \FOR{$w = 1, 2, \dots, 14$}
            \STATE{Call Algorithm \ref{alg:base} using $\mathcal{D}_{w}^{\EDGE}, \gamma^{\EDGE}$, and update $\theta_{w}^{\EDGE}$}
            \STATE{Set $A_{w}^{\EDGE} = \argmax_{\alpha} \langle \phi^{\EDGE}(S_{w}^{\EDGE}, \alpha), \theta_{w}^{\EDGE} \rangle$}
            \FOR{$d = 1, 2, \dots, 7$}
                \STATE{Call Algorithm \ref{alg:base} using $\mathcal{D}_{w,d}^{\CARE}, \gamma^{\CARE}$, and update $\theta_{w,d}^{\CARE}$}
                \STATE{Set $A_{w,d}^{\CARE} = \argmax_{\alpha} \langle \phi^{\CARE}(S_{w,d}^{\CARE}, \alpha), \theta_{w,d}^{\CARE} \rangle$}
                \FOR{$t = 1, 2$}
                    \STATE{Call Algorithm \ref{alg:base} using $\mathcal{D}_{w,d,t}^{\AYA}, \gamma^{\AYA}$, and update $\theta_{w,d,t}^{\AYA}$}
                    \STATE{$A_{w,d,t}^{\AYA} = \argmax_{\alpha} \langle \phi^{\AYA}(S_{w,d,t}^{\AYA}, \alpha), \theta_{w,d,t}^{\AYA} \rangle$}
                    \STATE{Environment generates $R_{w,d,t}^{\AYA}$ and next state $S_{w,d,t+1}$}
                    \STATE{Update $\mathcal{D}_{w,d,t}^{\AYA} = \mathcal{D}_{w,d,t-1}^{\AYA} \cup \{(S_{w,d,t}^{\AYA}, A_{w,d,t}^{\AYA}, R_{w,d,t}^{\AYA})\}$}
                \ENDFOR
                \STATE{Compute care-partner reward $\tilde{R}_{w,d}^{\CARE}$ based on Equ. (\ref{equ:care_rwd_app})} 
                \STATE{Update $\mathcal{D}_{w,d}^{\CARE} = \mathcal{D}_{w,d-1}^{\CARE} \cup \{(S_{w,d}^{\CARE}, A_{w,d}^{\CARE}, \tilde{R}_{w,d}^{\CARE})\}$}
            \ENDFOR
            \STATE{Compute relationship reward $\tilde{R}_{w}^{\EDGE}$ based on Equ. (\ref{equ:game_rwd_app})}
            \STATE{Update $\mathcal{D}_{w}^{\EDGE} = \mathcal{D}_{w-1}^{\EDGE} \cup \{(S_{w}^{\EDGE}, A_{w}^{\EDGE}, \tilde{R}_{w}^{\EDGE})\}$}
        \ENDFOR
    \end{algorithmic}
    \label{alg:multi_agent_surrogate}
\end{algorithm}

\begin{table}[hpt]
    \centering
    \caption{Prior mean for coefficients in the surrogate reward functions.}
    \label{tab:prior}
    \begin{tabular}{c|c|c|c|c|c}
    \toprule
    Agent & Intercept & $Y_w^{\EDGE}$ & $B_w^{\AYA}$  & $A_w^{\EDGE}$ & $A_w^{\EDGE} \cdot Y_w^{\EDGE}$ \\
    \midrule
    $\vbeta^{\EDGE}$ & $1$ & $1$ & $-1$ & $-1$ & $0.5$ \\
    \midrule
    \midrule
    Agent & Intercept & $Y_{w,d}^{\CARE}$ & $B_{w,d}^{\CARE}$ & $Y_{w-1}^{\EDGE}$ & $A_{w,d}^{\CARE}$  \\
    \midrule
    $\vbeta^{\CARE}$ & $1$ & $-1$ & $-1$ & $1$ & $-0.5$  \\
    \bottomrule
    \end{tabular}
    \end{table}
    
\section{The Dyadic Environment}
\label{app:testbed}

\subsection{Overview of the Simulated Dyadic Environment}


We construct a dyadic simulation environment to evaluate the performance of the proposed algorithm. The 1st order goal of the environment design is to replicate the noise level and structure that we expect to encounter in the forthcoming ADAPTS-HCT clinical trial. This noise often encompasses the stochasticity in the state transition of each participant and the heterogeneity across participants.

The environment is based on Roadmap 2.0, a care partner-facing mobile health application that provides daily positive psychology interventions to the care partner only. Roadmap 2.0 involves 171 dyads, each consisting of a patient undergone HCT (target person) and a care partner. Each participant in the dyad had the Roadmap mobile app on their smartphone and wore a Fitbit wrist tracker. The Fitbit wrist tracker recorded physical activity, heart rate, and sleep patterns. Furthermore, each participant was asked to self-report their mood via the Roadmap app every evening. A list of variables in Roadmap 2.0 is reported in Table \ref{tab:roadmap_variable}.

Roadmap 2.0 data is suitable for constructing a dyadic environment for developing the RL algorithm for ADAPTS-HCT in that Roadmap 2.0 has the same dyadic structure about the participants--post-HCT cancer patients and their care partner. Moreover, Roadmap 2.0 encompasses some context variables that align with those to be collected in ADAPTS-HCT, for example, the daily self-reported mood score.

\subsubsection{Overcoming impoverishment.} From the viewpoint of evaluating dyadic RL algorithms, this data is impoverished \cite{trella2022designing} mainly in two aspects. First, Roadmap 2.0 does not include micro-randomized daily or weekly intervention actions (i.e., whether to send a positive psychology message to the patient/care partner and whether to engage the dyad into a weekly game). Second, it does not include observations on the adherence to the medication--the primary reward signal, as well as other important measurements such as the strength of relationship quality. 

To overcome this impoverishment, we construct surrogate variables from the Roadmap 2.0 data to represent the variables intended to be collected in ADAPTS-HCT. A list of substitutes is reported in Table \ref{tab:roadmap_substitutes}. Worthnoting, the substitute for the AYA medication adherence is based on the step count. There is evidence on the association between the step count and the adherence. 

We further impute the treatment effects of the intervention actions so the marginal effects after normalization, which we call the standardized treatment effects (STE), are around 0.15, 0.3, and 0.5, corresponding to small, medium, and large effect sizes in typical behavioral science studies.

\subsubsection{Constructing the dyadic environment.} We follow the environment design in \cite{li2023dyadic}, which also uses the Roadmap 2.0 data, but primarily focuses on AYA intervention and relationship intervention. We extend the environment to include the care partner intervention. Specifically, we fit a separate multi-variate linear model for each participant in the dataset with the AR(1) working correlation using the generalized estimating equation (GEE) approach \cite{ziegler2010generalized,hojsgaard2006r}. We impute the treatment effects of the intervention actions based on the typical STE around 0.15, 0.3, and 0.5, which completes a generative model for the state transitions. The environment simulates a trial by randomly sampling dyads from the dataset, and simulate their trajectories based on the actions selected by the RL algorithm. The environment details are described in Appendix \ref{app:testbed}. Our experiments primarily focus on the three vanilla testbeds corresponding to the three STEs.


\subsection{Using the Roadmap 2.0 Dataset}

This section outlines our approach to addressing the limitations of the Roadmap 2.0 dataset, specifically its absence of micro-randomized interventions and reward signals.

To circumvent the lack of interventions, we impute treatment effects that represent the burden of the digital interventions, assuming that frequent notifications diminish both weekly and the daily treatment effects. Based on prior literature, we choose the scale of the treatment effect to be smaller than the baseline effect of features \cite{box1987empirical}. 

To address the missing reward signals, we use directly measurable variables in Roadmap 2.0 dataset as proxies to the outcomes we will observe in the real clinical trial. We approximate AYA adherence, $R_{w,d,t}^{\AYA}$, using the 12-hourly step count from Roadmap 2.0. Previous work has found the two values to be strongly correlated. Since adherence is a binary signal in the ADAPTS-HCT trial, we discretize step count into a binary variable. Furthermore, we approximate the carepartner's daily psychological distress, $Y_d^{\CARE}$, using the daily length of their sleep. Finally, the weekly relationship between the AYA and their carepartner is estimated using the self-reported mood as a surrogate. Specifically, we let  $Y_w^{\EDGE} = \mathbbm{1}\{\sum_{d = 1}^{7} \Mood_{w,d}^{\AYA} \geq \Mood^{\AYA}\} \mathbbm{1}\{\sum_{d = 1}^{7}\Mood_{w,d}^{\CARE} \geq \Mood^{\CARE}\}$. Here, $\Mood_{w,d}^{\AYA}$ is the daily self-reported mood on week $w$ and day $d$, and $\Mood^{\AYA}$ is the $q$-th quantile of the the weekly summed mood across all AYA observations. We choose the quantile level $q$ such that approximately 50\% of the dataset satisfies $Y_{w}^{\EDGE} = 1$.

Table \ref{tab:roadmap_substitutes} summarizes the main variables and their replacements from the Roadmap 2.0 dataset.  

\begin{table}[hpt]
    \centering
    \caption{Substitutes of the main variables from Roadmap 2.0 dataset.}
\resizebox{\textwidth}{!}{%
    \begin{tabular}{c|c}
        \hline
        Variables & Substitutes \\
        \hline
        \hline
        AYA adherence  & Binary step count $\mathbbm{1}\{\texttt{Step}_{w, d,t}^{\AYA} \geq \texttt{Step}^{\AYA}\}$  
        \\
        Carepartner distress & Carepartner daily length of sleep $\texttt{Sleep}_{w, d}^{\CARE}$ \\
        Weekly relationship quality & Mood indicator: $\mathbbm{1}_{\{\sum_{d} \texttt{Mood}_{w, d}^{\CARE} \geq \texttt{Mood}^{\CARE}\}} \mathbbm{1}_{\{\sum_{d} \texttt{Mood}_{w, d}^{\AYA} \geq \texttt{Mood}^{\AYA}\}}$ \\
        Effects of interventions $A_{w,d,t}^{\AYA}, A_{w,d}^{\CARE}$, $A_{w}^{\EDGE}$ & Imputed based on domain knowledge \\
        Effects of digital interventions burden $B_{w,d,t}^{\AYA}$, $B_{w,d}^{\CARE}$ & Imputed based on domain knowledge\\
        \hline
    \end{tabular}%
    }    \label{tab:roadmap_substitutes}
\end{table}

\begin{table}[hpt]
    \centering
    \caption{List of variables in Roadmap 2.0 and the measuring frequencies.}
    \begin{tabular}{c}
    \hline
    Variables\footnote{Note that all the variables are measured the same for the target person and carepartner.} \\
    \hline
    \hline
     $\texttt{Step}_{w, d, t}$: twice-daily cumulative step count\\
     $\texttt{Heart}_{w, d, t}$: twice-daily average heart rate\\
     $\texttt{Sleep}_{w, d}$: daily length of sleep\\
     $\texttt{Mood}_{w, d}$: daily self-report mood measurement\\
     \hline
    \end{tabular}
    \label{tab:roadmap_variable}
\end{table}

\subsection{Environment Model Design}

We now describe how these surrogate variables are used to build the full environment model. Our approach involves fitting two state transition models for digital intervention burden (AYA and carepartner) and three models for rewards (AYA adherence, carepartner stress, and relationship quality).

For all transition models, we fit the baseline parameters -- which represent system dynamics under no intervention --  for each dyad using its respective dataset and a generalized estimating equation \cite{hojsgaard2006r} approach. We impute the remaining parameters  using domain knowledge. Further detail on the choice of the coefficients is in Appendix \ref{sec:select_sim_params}. 

\textbf{Transition models for the AYA component: } The digital intervention burden transition for AYA follows a linear model with covariates $(B_{w,d,t}^{\AYA}, A_{w,d,t}^{\AYA}, A_{w}^{\EDGE})$.
\begin{align}
\label{equ:B_transition_AYA}
    B_{w,d,t+1}^{\AYA} \sim \theta^{\AYA}_{0} + \theta_{1}^{\AYA} B_{w,d,t}^{\AYA} + \theta_{2}^{\AYA} A_{w,d,t}^{\AYA} + \theta_{3}^{\AYA} A_{w}^{\EDGE} + \eta_{w,d,t}^{\AYA}, \nonumber\\
    \text{ where $\eta_{w,d,t}^{\AYA} \sim \mathcal{N}(0, (\omega^{\AYA})^2)$.} 
\end{align}

For the primary outcome, AYA adherence, we fit a generalized linear model with a sigmoid link function:

\begin{align}
R_{w,d,t}^{\AYA} &\sim \text{Bernoulli}(\text{sigmoid}(P_{w,d,t}^{\AYA})), \nonumber \\
P_{w,d,t}^{\AYA} &= (1-M_t)\big(\beta_{0, \AM}^{\AYA} + \beta^{\AYA}_{1, \AM} R_{w,d,t-1}^{\AYA} 
+ \beta_{2,\AM}^{\AYA} Y_{w-1}^{\EDGE} 
+ \beta_{3,\AM}^{\AYA} Y_{w,d-1}^{\CARE} + \beta_{4, \AM}^{\AYA} B_{w,d,t}^{\AYA} \nonumber \\
&\quad + \tau_{0, \AM}^{\AYA} A_{w,d,t}^{\AYA} 
+ \tau_{1, \AM}^{\AYA} A_{w,d,t}^{\AYA} Y_{w-1}^{\EDGE} 
+ \tau_{2, \AM}^{\AYA} A_{w,d,t}^{\AYA} B_{w,d,t}^{\AYA}\big) \nonumber \\
&\quad + M_t\big(\beta_{0, \PM}^{\AYA} + \beta^{\AYA}_{1, \PM} R_{w,d,t-1}^{\AYA} 
+ \beta_{2,\PM}^{\AYA} Y_{w-1}^{\EDGE} 
+ \beta_{3,\PM}^{\AYA} Y_{w,d-1,t}^{\CARE} + \beta_{4, \PM}^{\AYA} B_{w,d,t}^{\AYA} \nonumber \\
&\quad + \tau_{0, \PM}^{\AYA} A_{w,d,t}^{\AYA}  
+ \tau_{1, \PM}^{\AYA} A_{w,d,t}^{\AYA} Y_{w-1}^{\EDGE} 
+ \tau_{2, \PM}^{\AYA} A_{w,d,t}^{\AYA} B_{w,d,t}^{\AYA}\big)
\label{equ:R_Transition_AYA}
\end{align}

where $M_t$ is a decision window indicator defined as:

$$
    M_t = \left\{
    \begin{array}{clll}
         0 & \text{ if } t = 2k - 1 & (\text{AM decision window}) & \text{ for } k = 1, 2, \dots  \\
         1 & \text{ if } t = 2k & (\text{PM decision window}) &\text{ for } k = 1, 2, \dots
    \end{array},\right.
$$ 
Note that we exclude any effect of relationship interventions on AYA adherence as the game is designed without reinforcements and, thus, is not supposed to directly improve adherence.

\textbf{Transition models for the carepartner component: } The digital intervention burden transition for the carepartner is a linear model:

\begin{align}
\label{equ:B_transition_care}
    B_{w,d+1}^{\CARE} = \theta^{\CARE}_{0} + \theta_{1}^{\CARE} B_{w,d}^{\CARE} + \theta_{2}^{\CARE} A_{w,d}^{\CARE} + \theta_{3}^{\CARE} A_{w}^{\EDGE} + \eta_{w,d}^{\CARE}, \nonumber\\
    \text{ where $\eta_{w,d}^{\CARE} \sim \mathcal{N}(0, (\omega^{\CARE})^2)$.}
\end{align}

For the carepartner's psychological distress level, $R^{\CARE}_d$, we fit another linear model:
\begin{align}
    Y_{w,d}^{\CARE} = 
    &\beta_{0}^{\CARE} + \beta_{1}^{\CARE} Y_{w,d-1}^{\CARE} + \beta_{2}^{\CARE} R_{w,d,t-1}^{\AYA}  + 
    \beta_{3}^{\CARE} Y_{w-1}^{\EDGE} + \beta_{4}^{\CARE} B_{w,d}^{\CARE} + \nonumber \\
    &\quad \tau_{0}^{\CARE} A_{w,d}^{\CARE} +  
    \tau_{1}^{\CARE} A_{w,d}^{\CARE} Y_{w-1}^{\EDGE} +   
    \tau_{2}^{\CARE} A_{w,d}^{\CARE} B_{w,d}^{\CARE}  + \epsilon_{w,d}^{\CARE} \label{equ:R_transition_care}
\end{align}
where $\epsilon_{w,d}^{\CARE} \sim \mathcal{N}(0, (\sigma^{\CARE})^2)$.  Similar to (\ref{equ:R_Transition_AYA}), we do not include relationship intervention $A_{w-1}^{\EDGE}$.

\textbf{Transition model for the weekly relationship: } For the shared component, we only fit a transition model for the reward, which is the weekly relationship quality. Specifically, we fit a generalized linear model with a sigmoid link function: 

\begin{align}
Y_{w+1}^{\EDGE} \sim \text{Bernoulli}(\text{sigmoid}\left( \beta_{0}^{\EDGE} + \beta_{1}^{\EDGE}Y_{w}^{\EDGE}  + \beta_{2}^{\EDGE} \bar{R}_{w}^{\AYA} + \beta_{3}^{\EDGE} \bar{R}_{w}^{\CARE} \right. \nonumber \\
\left. + \tau_0^{\EDGE} A_{w}^{\EDGE} + \tau_1^{\EDGE} A_{w}^{\EDGE} (B_{w,d}^{\CARE} + B_{w,d,t}^{\AYA}))\right)
\label{equ:R_transition_rel}
\end{align}

where $\bar{R}_{w}^{\AYA} = \sum_{d=1}^{7} \sum_{t=1}^{2} \gamma^{14 - (7(w-1) + d) + 2(t-1)} R_{w,d,t}^{\AYA}$ is the exponentially weighted average of adherence within week $w$, and $\bar{R}_{w}^{\CARE} = \sum_{d=1}^{7} \gamma^{7-d} Y_{w,d}^{\CARE}$ is the exponentially weighted average of carepartner distress within week $w$. 

\subsection{Selecting Environment Model Parameters}
\label{sec:select_sim_params}

We list all the parameters that must be either imputed based on domain knowledge or estimated from the existing dataset. 

\begin{enumerate}
\item The baseline transition parameters $\beta$'s can be estimated directly from the dataset:
    \begin{enumerate}
        \item AYA state transition: $\vbeta^{\AYA}_{\AM} = (\beta_{0, \AM}^{\AYA}, \beta_{1, \AM}^{\AYA}, \beta_{2, \AM}^{\AYA}, \beta_{3, \AM}^{\AYA}, \beta_{4, \AM}^{\AYA})$ and $\vbeta^{\AYA}_{\PM} = (\beta_{0, \PM}^{\AYA}, \beta_{1, \PM}^{\AYA}, \beta_{2, \PM}^{\AYA}, \beta_{3, \PM}^{\AYA}, \beta_{4, \PM}^{\AYA})$.
        \item Carepartner state transition: $\vbeta^{\CARE} = (\beta_{0}^{\CARE}, \beta_{1}^{\CARE})$.
        \item Relationship transition: $\vbeta^{\EDGE} = (\beta_{0}^{\EDGE}, \beta_{1}^{\EDGE}, \beta_{2}^{\EDGE}, \beta_{3}^{\EDGE})$.
    \end{enumerate}
\item Imputed or tuned based on domain knowledge:
    \begin{enumerate}
        \item Burden transitions: coefficients $\boldsymbol{\theta}^{\AYA} = (\theta^{\AYA}_{0}, \theta^{\AYA}_{1}, \theta^{\AYA}_{2}, \theta^{\AYA}_{3})$, $\boldsymbol{\theta}^{\CARE} = (\theta^{\CARE}_{0}, \theta^{\CARE}_{1}, \theta^{\CARE}_{2}, \theta^{\CARE}_{3})$; burden noise variance $\omega^{\AYA}$ and $\omega^{\CARE}$.
        \item Main effects of burden: $\beta_{4, \AM}^{\AYA}, \beta_{4, \PM}^{\AYA}$, and $\beta_{4}^{\CARE}$.
        \item AYA treatment effects: $\{\tau_{i, \AM}^{\AYA}\}_{i = 0}^{2}$, $\{\tau_{i, \PM}^{\AYA}\}_{i = 0}^{2}$ and $\{\sigma_{i, \AM}^{\AYA}\}_{0 = 1}^{2}$, $\{\sigma_{i, \PM}^{\AYA}\}_{i = 0}^{2}$.
        \item Carepartner treatment effects: $\{\tau_{i}^{\CARE}\}_{i = 0}^{2}$ and $\{\sigma_{i}^{\CARE}\}_{i = 0}^{2}$.
        \item Relationship treatment effects: $\tau^{\EDGE}$ and $\sigma^{\EDGE}$.
    \end{enumerate}
\end{enumerate}

\textbf{Fitting parameters (1a-d):} We estimate the baseline transition parameters under no intervention directly from the Roadmap 2.0 dataset. For the parameters in Equation (\ref{equ:R_Transition_AYA}), we have the correspondences 
$\beta_{i, \AM}^{\AYA} = \hat{\beta}_{i, \AM}^{\AYA}$ and $\beta_{i, \PM}^{\AYA} = \hat{\beta}_{i, \PM}^{\AYA}$ for $i = 0, 1, \dots, 3$, where $\hat{\beta}_{i, \AM}^{\AYA}$ and $\hat{\beta}_{i, \PM}^{\AYA}$ are fitted coefficients obtained using the generalized estimating equation (GEE) approach. Since we assume that app burden only moderates the effects of AYA interventions without directly influencing adherence, we set $\beta_{4, \AM}^{\AYA} = \beta_{4, \PM}^{\AYA} = 0$. Similarly, for parameters in Equation (\ref{equ:R_transition_care}), the correspondence is $\beta_{i}^{\CARE} = \hat{\beta}_{i}^{\CARE}$ for $i = 0, \dots, 3$, and we set $\beta_{4}^{\CARE} = 0$ under the same assumption for carepartner distress. For the relationship quality model in Equation (\ref{equ:R_transition_rel}), the correspondence is $\beta_{i}^{\EDGE} = \hat{\beta}_{i}^{\EDGE}$ for $i = 0, \dots, 3$. Based on domain knowlege, we also truncate the parameters as follows: $\beta_{2, *}^{\AYA} = \max\{0, \hat{\beta}_{2, *}^{\AYA}\}$, reflecting the assumption that weekly relationship quality non-negatively influences AYA adherence, $\beta_{3, *}^{\AYA} = \min\{0, \hat{\beta}_{3, *}^{\AYA}\}$, as carepartner distress is expected to negatively influence adherence, and $\beta_{3}^{\EDGE} = \min\{0, \hat{\beta}_{3}^{\EDGE}\}$ as carepartner distress is expected negatively impact relationship quality.



\textbf{Imputing Burden Transitions (2a):}
We set $
  \theta_{1}^{\AYA} = \tfrac{13}{14}, \quad \theta_{1}^{\CARE} = \tfrac{6}{7},
$ so that the memory of digital burden spans roughly one week for both AYA and carepartner. We choose
$\theta_{2}^{\AYA} = 5\,\theta_{3}^{\AYA} = 1, 
  \quad
  \theta_{2}^{\CARE} = 5\,\theta_{3}^{\CARE} = 1$ so that daily interventions exert five times more burden than the weekly relationship intervention. The intercepts are $
  \theta_{0}^{\AYA} = 0.2, 
  \quad
  \theta_{0}^{\CARE} = 0.2$, and chosen so that participants have around a 20\% baseline burden even without an intervention. We set
$
  \omega^{\AYA} = \omega^{\CARE} = 2.4
$ to obtain a moderate noise-to-signal ratio, set so that 
\(
    (\theta_{1}^{\AYA} + \theta_{2}^{\AYA}) / \omega^{\AYA} 
    \approx 0.5
\).
We then truncate burdens at zero and standardize them separately for AYA and carepartner by simulating 10,000 steps with random interventions.

\textbf{Imputing main effects of app burden (2b).} We set $\beta_{4, \AM}^{\AYA} = \beta_{4, \PM}^{\AYA} = \beta_{4}^{\CARE} = 0$ based on the assumption that digital app intervention burden does not directly affect AYA adherence or carepartner distress, unless through moderating the digital interventions.

\textbf{Imputing treatment Effects (2c--2f):}
Since digital health environments are noisy, treatment terms likely have a lower effect on transitions than the baseline transitions under no intervention. Hence, we scale all intervention effects relative to the baseline effects using a single, global hyperparameter $C_{\text{treat}}$. 

For each time of the day (AM or PM), the AYA intervention increases adherence by $\tau_{0, *}^{\AYA} = C_{\text{treat}} \bigl|\beta_{1, *}^{\AYA}\bigr|$, where $* \in \{\mathrm{AM}, \mathrm{PM}\}$ and $\beta_{1, *}$ is the corresponding baseline coefficient estimated from Roadmap 2.0. 

We further define $\bigl|\beta_{1, *}^{\AYA}\bigr|$ and $\tau_{\text{burden}, *}^{\AYA} = -C_{\text{treat}} \bigl|\beta_{1, *}^{\AYA}\bigr|$ because the AYA intervention's effectiveness can be increased by good relationship quality and decreased by high digital-intervention burden.

To account for individual heterogeneity across dyads, each treatment-effect coefficient has an associated random effect with variance $\sigma_{0, *}^{\AYA} = C_{\text{treat}} \sigma_{\beta_{1, *}^{\AYA}}$, where $\sigma_{\beta_{1, *}^{\AYA}}$ is the empirical standard deviation across dyads of the baseline coefficient $\beta_{1, *}^{\AYA}$. 

For carepartner interventions, the main effect on distress is scaled as $\tau_{0}^{\CARE} = -C_{\text{treat}} \bigl|\beta_{1}^{\CARE}\bigr|$, where the negative sign is due to the intervention reducing distress. Lastly, the effect of the weekly relationship intervention on improving relationship quality is given by $\tau^{\EDGE} = C_{\text{treat}} \bigl|\beta_{1}^{\EDGE}\bigr|$.

We summarize the imputation design in Table \ref{tab:imputation}. 

\begin{table}[hpt]
    \centering
    \caption{Summary of burden transition design and treatment effects design.}
    \begin{tabular}{c|>{\centering\arraybackslash}p{0.4\textwidth}} 
    \hline
    \multicolumn{2}{c}{Burden transition}\\
    \hline
    \hline
    Intercept $\theta_{0}^{\AYA}$  & Based on domain knowledge $\theta_{0}^{\AYA} = 0.2$ \\
    Intervention coefficients $\theta_2^{\AYA}, \theta_3^{\AYA}$ & $\theta_2^{\AYA} = 5\theta^{\AYA}_{3} = 1$ (Because relationship intervention produces lower burden) \\
    Noise standard deviation $\omega^{\AYA}$ & Based on the typical noise-to-signal ratio $\omega^{\AYA} = 2.4$ \\
    \hline
    \multicolumn{2}{c}{Treatment effect for twice-daily adherence transition (* stands for AM or PM)}\\
    \hline
    \hline
    Main effect of AYA intervention $\tau_{0, *}^{\AYA}$ & Hyper-parameter $\tau_{0, *}^{\AYA} = C_{\Treat}|\beta_{1, *}^{\AYA}| $ \\
    Rel. and AYA int. interaction $\tau_{2, *}^{\AYA}$ & Hyper-parameter $\tau_{1, *}^{\AYA} = C_{\Treat}|\beta_{1, *}^{\AYA}|$ \\
    Burden and AYA int. interaction $\tau_{4, *}^{\AYA}$ & Hyper-parameter $\tau_{2, *}^{\AYA} = C_{\Treat}|\beta_{1, *}^{\AYA}|$ \\
    Random treatment variance $\{\sigma_{i, *}^{\AYA}\}_{i = 0}^5$ & Scales with the variance of $\beta_{1, *}^{\AYA}$: $\sigma_{i, *}^{\AYA} = \tau_{i, *}^{\AYA} \cdot  \sigma_{\beta_{1, *}^{\AYA}} / |\beta_{1, *}^{\AYA}|$ \\
    \hline
    \multicolumn{2}{c}{Treatment effect for weekly relationship transition}\\
    \hline
    \hline
    Main effect of relationship int. $\tau^{\EDGE}$ & Hyper-parameter $\tau^{\EDGE} = C_{\Treat}|\beta_1^{\EDGE}|$  \\
    \hline
    \end{tabular}
    \label{tab:imputation}
\end{table}

\textbf{Tuning $C_{\Treat}$: } We tune the hyperparameter  $C_{\Treat}$ such that the standardized treatment effects (STE) are around 0.15, 0.3, and 0.5, where STE is defined as:
\begin{equation}
    \operatorname{STE}(C_{\Treat}) = \frac{\mathbb{E}\left[\mathbb{E}[\text{CR}(\pi^*_{e}) \mid e] - \mathbb{E}[\text{CR}(\pi_0, e) \mid e]\right]}{\sqrt{\operatorname{Var}(\mathbb{E}[\text{CR}(\pi_0, e) \mid e])}},
    \label{equ:STE}
\end{equation}
Here, $e$ corresponds to the resulting environment model for dyad $e$ when the hyperparameter is set to be $C_{\Treat}$, and $\pi_e^*$ is the optimal policy for dyad $e$. $\text{CR}(\pi, e)$ is the cumulative rewards earned by running policy $\pi$ on dyad $e$, and $\pi_0$ is the reference policy that always chooses action 0 for all components. 

\begin{figure}[hpt]
    \centering
    \includegraphics[width=0.8\textwidth]{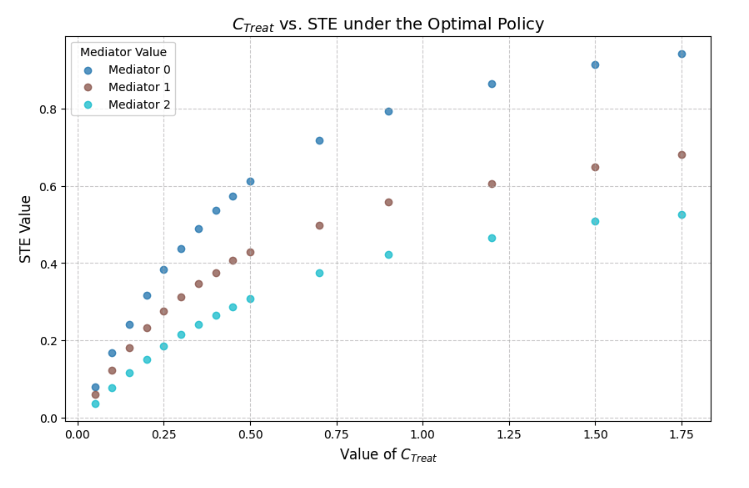}
    \caption{Relationship between the hyperparameters and the STE, categorized by the mediator effect value.}
    \label{fig:ste}
\end{figure}

Figure \ref{fig:ste} plots the value of the hyperparameter versus the STE computed using the optimal policy in the environment defined by the hyperparameter. We outline our approximation of the optimal policy in Appendix \ref{sec:optpol}. By default, we choose an environment with mediator effect = 1. This results in three dyadic environments, which we summarize in Table \ref{tab:test-bed}.

\begin{table}[ht]
    \centering
    \caption{Summary of all testbeds}
    \begin{tabular}{c|c}
     Treatment effect size & Value of $C_{\Treat}$ \\
    \hline
       0.15 (Small)  & 0.2 \\
       0.3 (Medium) & 0.3 \\
       0.5 (Large) & 0.5 \\
    \end{tabular}
    \label{tab:test-bed}
\end{table}

\subsection{Optimal Policy Approximation}
\label{sec:optpol}


To approximate the optimal policy, we generate a dataset under a random policy with $P(A_{w,d,t}^{\AYA}=1)= P(A_{w,d}^{\CARE}=1) = P(A_w^{\EDGE}=1)=0.5$ and apply offline Q-learning on this dataset. To make the computation tractable, we discretize and subset the features. Specifically, we use six features: the intercept, AYA adherence, carepartner distress, AYA burden, carepartner burden, and relationship quality. Numerical features (carepartner distress, AYA burden, and carepartner burden) are discretized into 10 bins.

Finally, we evaluate the performance of this approximation against other baseline policies, including micro-randomized actions with fixed probabilities of 0.5, 0.6, 0.7, 0.8, and 0.9. Our approximation consistently outperforms these baselines.

\subsection{Evidence of the Need for Collaboration in the Dyadic Environment}
\label{app:evidence-collab}

To show that each agent impacts the performance of other agents, we consider the following toy setting.
We fix the care partner agent's randomization probability at 0.5 and vary the AYA agent’s probability to be 0.25 and 0.75. Then, for each fixed AYA agent's probability, we identify the value of the relationship agent’s probability that maximizes average weekly adherence. We find that this relationship probability changes from $1.0$ to $0.0$ when we change AYA agent's probability from 0.25 to 0.75. 

We repeat this experiment for the care partner agent by fixing the AYA agent’s probability at 0.5 and varying the relationship agent’s probability to be 0.25 and 0.75. Similarly, we find that the care partner agent's probability that maximizes adherence changes from 0.6 to 0.5 when we vary the relationship probability from 0.25 to 0.75.

These results indicate that the agents must change their behavior to account for the other agents' behavior.




\section{Additional Results}

\label{app:additional_results}
\subsection{Ablation Study}

\begin{figure}[hpt]
    \centering
    \begin{subfigure}[b]{0.31\textwidth}
        \includegraphics[width=1\textwidth]{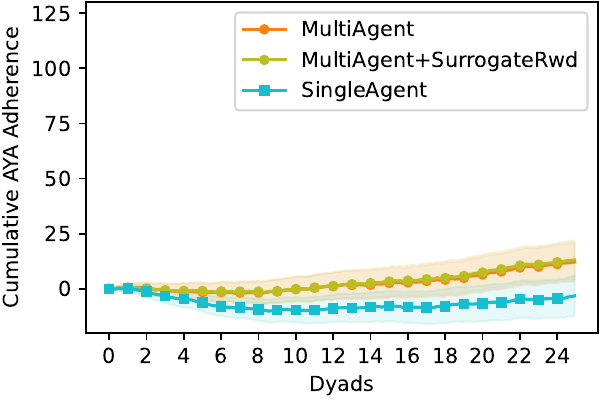}
        \caption{STE 0.15}
    \end{subfigure}
    \begin{subfigure}[b]{0.31\textwidth}
        \includegraphics[width=1\textwidth]{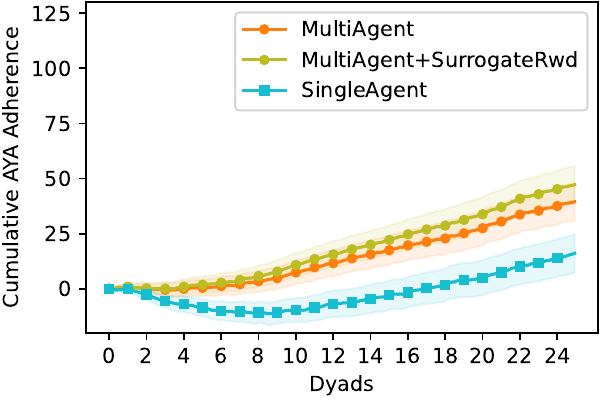}
        \caption{STE 0.3}
    \end{subfigure}
    \begin{subfigure}[b]{0.31\textwidth}
        \includegraphics[width=1\textwidth]{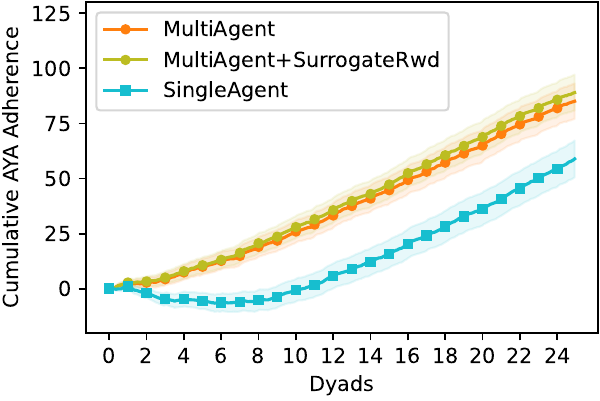}
        \caption{STE 0.5}
    \end{subfigure}
    \caption{Cumulative rewards improvement over the uniform random policy for all three components under the testbed without the effect of care-partner distress onto relationship quality or the effect of relationship quality onto AYA's adherence.}
    \label{fig:mediator0}
\end{figure}

\paragraph{No Mediator Effect} The improvement from using a surrogate reward is through the effects of the mediator variables. For example, the relationship intervention $A_{w}^{\EDGE}$ improves the mediator relationship, which may improve the primary outcome, medication adherence. The care-partner intervention $A_{w,d}^{\CARE}$ mitigates the distress, which may improve relationship. In Fig. \ref{fig:mediator0}, we run the all three algorithms under a testbed variant for which we force the above two mediator effects to be 0, i.e., no effect from relationship to adherence or effect from distress to relationship. In this testbed variant, \texttt{MutiAgent+SurrogateRwd} performs the same as \texttt{MutiAgent}--there is no cost of reward learning under no mediator effect.

\paragraph{Other Testbed Variants.} To further violate the assumptions made from the causal diagram, we made the following two changes to test the robustness of our proposed algorithm: 1) we add a direct effect from care-partner psychological distress to AYA medication adherence; 2) we generate random mediator effects, effect from relationship to adherence and effect from distress to relationship. This later one violates the monotonicity assumptions learned from principles.

\subsection{Collaboration of Multi-Agent RL} 

We train each individual agent in the \texttt{MultiAgent+SurrogateRwd} algorithm over 1000 dyads under the STE 0.5 environment, while fixing the randomization probability of the other agents. We denote the randomization probability of the AYA agent, care partner agent, and relationship agent as $p^{\AYA}$, $p^{\CARE}$, and $p^{\EDGE}$ respectively.

We first train the relationship agent while fixing $p^{\CARE} = 0.5$. We see that the average probability of sending an intervention for the relationship agent is 0.57 and 0.42 under $p^{\AYA} = 0.25$ and $0.75$, respectively. This indicates that the relationship agent learns to \textit{reduce} the intervention probability when the AYA agent is more likely to send an intervention. 

Similarly, we train the care partner agent while fixing $p^{\AYA} = 0.5$. We see that the average probability of sending an intervention for the care partner agent is 0.61 and 0.45 under $p^{\EDGE} = 0.25$ and $0.75$, respectively. This indicates that the care partner agent learns to \textit{reduce} the intervention probability when the relationship agent is more likely to send an intervention.
\end{document}